%% file: main.tex
\documentclass{article}

\usepackage{microtype}
\usepackage{graphicx}
\usepackage{subfigure}
\usepackage{booktabs} 

\usepackage{hyperref}

\usepackage[accepted]{icml2024}

\usepackage{amsmath}
\usepackage{amssymb}
\usepackage{mathtools}
\usepackage{amsthm}

\usepackage[capitalize,noabbrev]{cleveref}

\theoremstyle{plain}

\theoremstyle{definition}

\theoremstyle{remark}

\usepackage{multirow}

\usepackage[textsize=tiny]{todonotes}

\icmltitlerunning{Efficient Non-parametric Uncertainty Quantification for Black-box Large Language Models and Decision Planning}

\begin{document}

\twocolumn[
\icmltitle{Efficient Non-Parametric Uncertainty Quantification for Black-Box Large Language Models and Decision Planning}



\icmlsetsymbol{equal}{*}

\begin{icmlauthorlist}
\icmlauthor{Yao-Hung Hubert Tsai}{apple}
\icmlauthor{Walter Talbott}{apple}
\icmlauthor{Jian Zhang}{apple}
\end{icmlauthorlist}

\icmlaffiliation{apple}{Apple}

\icmlcorrespondingauthor{Jian Zhang}{jianz@apple.com}

\icmlkeywords{Uncertainty Quantification in Language Models, Planning}

\vskip 0.3in
]



\printAffiliationsAndNotice{}  

\begin{abstract}
Step-by-step decision planning with large language models (LLMs) is gaining attention in AI agent development. This paper focuses on decision planning with uncertainty estimation to address the hallucination problem in language models. Existing approaches are either white-box or computationally demanding, limiting use of black-box proprietary LLMs within budgets. The paper's first contribution is a non-parametric uncertainty quantification method for LLMs, efficiently estimating point-wise dependencies between input-decision on the fly with a single inference, without access to token logits. This estimator informs the statistical interpretation of decision trustworthiness. The second contribution outlines a systematic design for a decision-making agent, generating actions like ``turn on the bathroom light'' based on user prompts such as ``take a bath''. Users will be asked to provide preferences when more than one action has high estimated point-wise dependencies. In conclusion, our uncertainty estimation and decision-making agent design offer a cost-efficient approach for AI agent development.
\end{abstract}

\input{sections/intro}

\input{sections/related}

\input{sections/uncertainty}

\input{sections/smarthomeagent}

\input{sections/exp}

\input{sections/conclusion}

\nocite{langley00}

\bibliography{uncertainty}
\bibliographystyle{icml2024}


\end{document}

%% file: sections/intro.tex
\section{Introduction}
\label{sec:intro}

The key distinction between traditional software and AI agents lies in the AI agent's ability to make decisions based on human requests, often conveyed in natural language. For instance, when prompted to book a trip to Europe, the AI agent should consider the user's historical preferences and decide between destinations like Vienna or Prague. However, in cases where both options are equally suitable, the AI agent shall seek human input for clarification, allowing users to disambiguate the decision. This process underscores the importance of uncertainty quantification, validating the plausibility of one or more decisions. Furthermore, single-round decision-making is insufficient. AI agents must engage in multi-round decision-making, also known as step-by-step decision planning. For example, having chosen Prague as the destination, the AI agent then needs to decide on accommodation based on nearby tourist attractions. In this work, we delve into the intricacies of step-by-step decision-making by AI agents and explore the role of efficient uncertainty quantification methods in enhancing the decision-making process.

Due to the surge in artificial intelligence, large language models (LLMs)~\cite{touvron2023llama,touvron2023llama2,jiang2023mistral,jiang2024mixtral,openai2023gpt4} have demonstrated their potential to fuel step-by-step decision planning in AI applications across domains such as robotics~\cite{ahn2022can,huang2022inner,ren2023robots}, workflow automation~\cite{yang2023auto,wu2023autogen,hong2023metagpt}, and medicine~\cite{thirunavukarasu2023large}. While uncertainty quantification can mitigate the hallucination problem in LLMs (i.e., the tendency for LLMs to be overly confident in their output, whether the decision is correct or not), current approaches either necessitate access to token logits~\cite{fomicheva2020unsupervised,kuhn2023semantic,takayama2019relevant,van2022mutual,malinin2020uncertainty,lee2018simple,yoo2022detection,ren2022out,vazhentsev2022uncertainty,kadavath2022language} or involve high computational costs~\cite{lin2023generating}. Unfortunately, the requirements hampers the application of building AI agents with black-box proprietary LLMs within budget constraints, such as using GPT-4~\cite{openai2023gpt4} as the backend for the AI agent. To address this concern, we propose an elegant and efficient uncertainty quantification approach for step-by-step decisions in language models. Then, building on this approach, we showcase the systematic design of a lightweight yet powerful decision-making agent.

\input{fig_tex/illus}

Our uncertainty quantification approach utilizes point-wise dependency neural estimation~\cite{tsai2020neural} to estimate the point-wise dependency $\frac{p(x,y)}{p(x)p(y)}$ between $x$ (user prompt and AI agent's history decisions) and $y$ (AI agent's current decision). Notably, this estimation is efficient, achieved with a single inference from an auxiliary neural network. In contrast, current black-box approaches~\cite{lin2023generating} require the AI agent to generate multiple decisions to assess decision quality, i.e., $p(y|x)$. The point-wise dependency $\frac{p(x,y)}{p(x)p(y)}$ has a clear statistical interpretation: values greater than $1.0$ indicate correlation, while values lower than $1.0$ indicate negative correlation. Conformal prediction~\cite{shafer2008tutorial} can be further incorporated to establish a threshold for $\frac{p(x,y)}{p(x)p(y)}$, determining whether to trust the decision. Previous studies have demonstrated that conformal prediction is an effective tool for uncertainty quantification in language models~\cite{ren2023robots,quach2023conformal,kumar2023conformal}.

Utilizing the introduced uncertainty quantification approach, we devise a decision-making agent capable of responding to a user prompt such as ``writing in the study'' with a series of decisions, including 1) ``study room light: soft'', 2) ``study room music player: play quiet music'', and then 3) ``desk lamp: on''. Our system encompasses three stages: data collection, model training, and deployment. We provide an illustration in Figure~\ref{fig:illus}.

In the data collection stage, we gather $20k$ user prompts and corresponding potential home agent actions. Examples include \{user prompt: time to fix things; actions: garage light: on, tool drawer: open\} and \{user prompt: water the plants; actions: outdoor lights: on, outdoor speaker: play laid-back music, smart sprinkler: on\}. 

During the model training stage, we initially conduct instruction fine-tuning using Mistral-7B-Instruct-v0.1\footnote{We treat Mistral-7B-Instruct-v0.1 as a black-box model by not utilizing its token output logits. One can find the model in \url{https://huggingface.co/mistralai/Mistral-7B-Instruct-v0.1}.}. Note that we design the fine-tuning to generate all possible actions at once. Subsequently, we train an auxiliary neural network for point-wise dependency neural estimation between user prompts, history decisions, and current decisions.

The deployment stage employs an interactive process. We first use conformal prediction to establish the threshold for point-wise dependency from the calibration data split. Then, based on our request, we prompt the language model, which performs a one-shot inference, generating all possible actions. Our neural point-wise dependency estimator computes scores for all pairs of user prompts and generated actions. If only one score exceeds the threshold, the decision-making agent selects that action and appends it to the user prompt. If multiple scores surpasses the threshold, the decision-making agent asks the user to choose a preferred action, which is then appended to the user prompt. The decision-making agent ceases generating actions when none of the scores exceeded the threshold.

We assess the decision-making agent's effectiveness by examining the mean precision, recall, and F1 score of generated actions compared to true actions on the evaluation data. Our findings highlight that step-by-step decision planning outperforms all-at-once decision planning in terms of F1 score. Furthermore, increasing the threshold for estimated point-wise dependency during action selection enhances mean precision. The best performance is achieved by consistently selecting the action with the maximum estimated point-wise dependency in step-by-step decision planning, indicating a higher likelihood of it being the true action. 

%% file: fig_tex/illus.tex
\begin{figure*}[th!]
    \centering
    \includegraphics[width=1.0\linewidth]{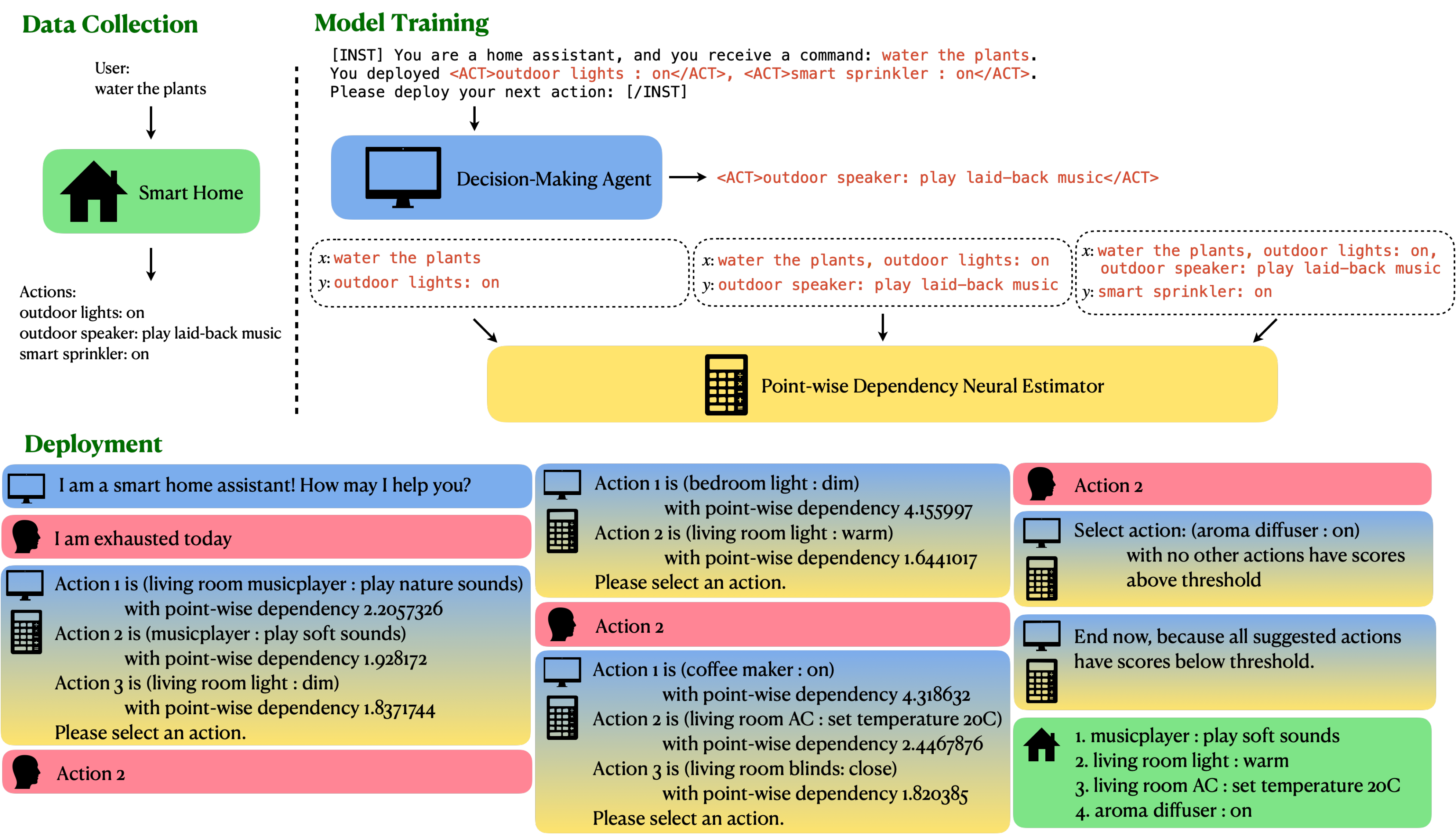}
    \vspace{-4mm}
    \caption{A decision-making agent design. During the {\em data collection} phase, we curate a dataset comprising 20,000 pairs associating user requests with smart home actions, recognizing the potential for multiple actions per request. In the subsequent {\em model training} stage, we conduct instruction fine-tuning. The agent, utilizing a robust language model, generates a comprehensive set of actions based on user requests and prior actions. Additionally, we train a point-wise dependency neural estimator, establishing relationships among user requests, historical actions, and the current action. Moving to the {\em deployment} stage, we integrate the decision-making agent with the neural estimator to enumerate potential actions guided by point-wise dependencies exceeding a threshold determined from calibration data. User interaction occurs when multiple actions are enumerated, prompting user selection. For a single enumerated action, the agent executes it directly, and in the absence of any enumerated actions, the agent ceases operation. We use green color to denote the smart home, blue color to denote the decision-making agent, yellow color to denote the point-wise dependency neural estimator, gradients of the blue and the yellow color to denote the combination between the agent and the neural estimator, and pink color to denote the user.}
    \label{fig:illus}
\end{figure*}

%% file: sections/related.tex
\section{Related Work}
\label{sec:related}

This paper covers a broad literature. In this section, our focus is specifically directed towards intersections between large language models (LLMs) with uncertainty quantification, AI agents, and decision planning.

{\bf Uncertainty Quantification.} Uncertainty quantification plays a crucial role in ensuring the safe and responsible utilization of large language models (LLMs). Its primary function lies in disambiguating over-confident yet incorrect responses from LLMs, a challenge commonly known as the hallucination problem~\cite{zhang2023siren}. However, the task of uncertainty quantification is inherently intricate~\cite{smith2013uncertainty,belghazi2018mutual}, and its complexity intensifies in the context of text generation using LLMs, particularly due to the nuances of sampling and beam searches~\cite{lin2023generating,takayama2019relevant,van2022mutual,kuhn2023semantic,malinin2020uncertainty,kadavath2022language}. Categorizing related approaches reveals two primary methodologies: the {\em white-box} and {\em black-box} methods.

On the one hand, the {\em white-box} approach necessitates access to logits, internal layer outputs, or the LLMs themselves. Some approaches involve forming the probability of output generation given user input~\cite{fomicheva2020unsupervised,kuhn2023semantic,takayama2019relevant,van2022mutual,malinin2020uncertainty}. Others focus on out-of-distribution detection~\cite{lee2018simple,yoo2022detection,ren2022out,vazhentsev2022uncertainty}, determining whether a generation aligns with the training distribution. And some approaches prompt LLMs to self-assess the correctness of their outputs~\cite{kadavath2022language}. 

On the other hand, the {\em black-box} approach tackles the challenge of utilizing proprietary large language models (LLMs) where only output texts are accessible. A method~\cite{lin2023generating} suggests estimating the probability of an output response by generating multiple responses corresponding to the input. This estimation involves assessing similarities between different responses, although we contend that defining similarity between responses introduces another challenge. The primary drawback of existing black-box approaches is their computational intensity, demanding multiple inferences. In contrast, our approach offers an elegant and efficient method for quantifying uncertainty in black-box LLMs. We eliminate the need for multiple generations given an input and the computation of sentence similarities.

{\bf Decision Planning.} Step-by-step decision planning utilizing language models has been demonstrated in various domains, including robotics~\cite{ren2023robots,brohan2023rt,driess2023palm,tsai2023multimodal,li2022pre} and self-driving~\cite{sha2023languagempc,zhou2023navgpt}. The overarching objective is to leverage LLMs for generating actions based on the current observation, historical observations, and goal specification. Contrasting with traditional decision-making approaches~\cite{chen2021decision}, the advantage of employing LLMs lies in their capacity to comprehend and reason with complex human languages.

However, the utilization of LLMs in decision planning introduces challenges, particularly the hallucination problem, which can potentially disrupt the decision-making process. An early incorrect decision, influenced by hallucinations, can have a cascading effect on subsequent decisions. In addition to relying on uncertainty quantification to address the hallucination problem, other researchers advocate for leveraging explicit~\cite{zhou2023navgpt} or implicit~\cite{tsai2023multimodal} memory modules, a process known as augmented retrieval~\cite{guu2020retrieval,izacard2022few}.

Furthermore, an emerging aspect in utilizing LLMs for decision planning focuses on enhancing inference efficiency~\cite{alizadeh2023llm,sun2020mobilebert}, enabling deployment on personal devices with mobile computing capabilities. This paper specifically covers certain facets of employing LLMs for step-by-step decision planning. Notably, we introduce an efficient uncertainty quantification method to mitigate hallucinations and outline the systematic design of a decision-making agent that incorporates past actions when making new decisions.

{\bf AI Agent.} The objective of an AI agent is to design and develop systems capable of emulating human-like intelligence and abilities~\cite{russell2010artificial,wooldridge1995intelligent}. The evolution of LLMs~\cite{openai2023gpt4} has significantly enhanced the performance of these agents~\cite{park2023generative,liu2023training,sumers2023cognitive,openai2023gpt4}, where LLMs serve as the cognitive center. Furthermore, the development goes beyond pure text to encompass diverse perceptual abilities and tool usage~\cite{nakano2021webgpt,yao2022react,schick2023toolformer,lu2023chameleon}. A pivotal aspect of an AI agent is its capacity to interact with humans~\cite{akyurek2023rl4f,peng2023check,liu2023languages}, enabling iterative improvement or collaborative co-piloting in production. Applications of such agents span across workflow automation~\cite{wu2023autogen}, software development~\cite{li2023camel,qian2023communicative}, scientific research~\cite{boiko2023emergent}, and medicine~\cite{dai2023ad}. This paper specifically focuses on the design of an AI agent tailored for a user-interactive decision-making experience.

%% file: sections/uncertainty.tex
\section{Efficient Non-parametric Uncertainty Quantification for Language Models}
\label{sec:uncertainty}

{\bf Notations.} We define the training, calibration, and evaluation data splits as ${\mathcal D}_{\rm train}$, ${\mathcal D}_{\rm calib}$, ${\mathcal D}_{\rm eval}$. Each data split consists of tuples between a user prompt $x$ and the corresponding actions $\{a^j\}_{j=1}^{n_x}$, where $n_x$ denotes the number of actions associated with the user prompt. For instance, $x$ is {\em water the plants} and $\{a^j\}$ are {\em \{outdoor lights: on, outdoor speaker: play laid-back music, smart sprinkler: on\}}. We denote their sample spaces as $X$ and $A$.

\subsection{Point-wise Dependency Neural Estimation} 

The objective of uncertainty quantification is to measure uncertainty, such as the conditional probability $p(a|x)$~\cite{fomicheva2020unsupervised}, representing the relationship between a pair of action $a$ and user prompt $x$. This paper employs techniques introduced in prior work~\cite{tsai2020neural}, utilizing a neural network $\theta$ to estimate the point-wise dependency $r(a,x)=\frac{p(a, x)}{p(a)p(x)}$, denoted as $r_\theta(a,x)$, between $a$ and $x$. The interpretation here is that a higher value of $r_\theta(a,x)$ indicates a stronger correlation between $a$ and $x$. Specifically, $r(a,x) < 1$ implies negative dependence between $a$ and $x$, $r(a,x) = 1$ suggests independence, and $r(a,x) > 1$ suggests positive dependence.

We adopt the relative predictive coding method~\cite{tsai2021self} as a stabilized version of the density-ratio fitting method~\cite{tsai2020neural}:
\begin{equation}
\begin{split}
\underset{\theta}{\rm sup}\,\, & \mathbb{E}_{P_{AX}} [r^*_\theta(a,x)] - \alpha \mathbb{E}_{P_AP_X} [r^*_\theta(a,x)] \\
& - \frac{\beta}{2} \mathbb{E}_{P_{AX}} [{r^*_\theta}^2(a,x)] - \frac{\gamma}{2} \mathbb{E}_{P_{A}P_{X}} [{r^*_\theta}^2(a,x)].
\end{split}
\label{eq:rpc}
\end{equation}

We denote the sampling of an associated pair of action and user prompt as $(a,x) \sim P_{AX}$, exemplified by {\em smart sprinkler: on} and {\em water the plants}. Likewise, $(a,x) \sim P_AP_X$ signifies the sampling of an unassociated pair of action and user prompt, such as {\em smart sprinkler: on} and {\em feeling hungry}. $\alpha, \beta, \gamma$ are hyper-parameters, and ${r_\theta}(a,x) = \frac{\gamma r^*_\theta(a,x) + \alpha}{ 1- \beta r^*_\theta(a,x)}$ in accordance with Lemma 1 in the prior work~\cite{tsai2021self}.

We select $\alpha=1.0, \beta=0.005, \gamma=0.1$ consistently throughout the paper, and the design of $r_\theta^*(a,x)$ is expressed as
$$
r_\theta^*(a,x) = \langle f_a^l\circ f_a^g \circ f^{llm} (a), f_x^l\circ f_x^g \circ f^{llm} (x) \rangle.
$$
Here, $\langle \cdot , \cdot \rangle$ denotes the inner product. $f^{llm}$ represents a pre-trained large language model (specifically, Mistral-7B-0.1~\cite{jiang2023mistral} is adopted). $f_a^g$/$f_x^g$ are gated recurrent units (GRUs)~\cite{cho2014learning} applied over the sequence outputs from $f^{llm}$, while $f_a^l$/$f_x^l$ are fully-connected layers attached from the last unit in $f_a^g$/$f_x^g$. The network $\theta$ exists in $f_a^g$/$f_x^g$ and $f_a^l$/$f_x^l$.

\subsection{Extension to Step-by-step Planning}

To establish a point-wise dependency neural estimator between an action and a user prompt, we can perform gradient updates on the objective~\eqref{eq:rpc}. Nonetheless, we also aim to understand the confidence in executing an action $a$ based on user prompt $x$ and the history of actions ${\bf a}_{\rm history}:=\{a_{-1}, a_{-2}, a_{-3}, \cdots\}$, where $\{a_{-1}, a_{-2}, a_{-3}, \cdots\}$ represent previously executed actions. 

To achieve this, we introduce $X'$ as the sample space that encompasses user prompts and any taken actions. In the example, $x' \sim X'$ could represent \{``{\em water the plants}''\}, \{``{\em water the plants, outdoor lights: on}''\}, or \{``{\em water the plants, outdoor lights: on, outdoor speaker: play laid-back music}''\}. Notably, we achieve this expansion by simply appending the taken actions to the user prompt for $X'$. Consequently, eq.~\eqref{eq:rpc} undergoes modification as follows:  
\begin{equation}
\begin{split}
\underset{\theta}{\rm sup}\,\, & \mathbb{E}_{P_{AX'}} [r^*_\theta(a,x')] - \alpha \mathbb{E}_{P_AP_{X'}} [r^*_\theta(a,x')] \\
& - \frac{\beta}{2} \mathbb{E}_{P_{AX'}} [{r^*_\theta}^2(a,x')] - \frac{\gamma}{2} \mathbb{E}_{P_{A}P_{X'}} [{r^*_\theta}^2(a,x')].
\end{split}
\label{eq:rpc2}
\end{equation}

The estimated point-wise dependency between an action $a$ and user prompt/history actions $x'$ is then ${r_\theta}(a,x') = \frac{\gamma r^*_\theta(a,x') + \alpha}{ 1- \beta r^*_\theta(a,x')}$.

%% file: sections/smarthomeagent.tex
\section{Decision-Making Agent}
\label{sec:smarthome}

In this section, we present the design of our decision-making AI agent, highlighting its capabilities in 1) controlling various functions within a synthetic smart home, 2) comprehending human requests, 3) executing actions in a step-by-step manner, and 4) soliciting human assistance when confronted with ambiguous outputs (e.g., multiple plausible actions), particularly when coupled with our uncertainty quantification technique~(see Section \ref{sec:uncertainty}). The design is implemented across three key stages: data collection, model training, and deployment.

\subsection{Data Collection}

We annotate $20k$ data with each data being user prompt and the associated actions of the smart home. Examples are the following:
\begin{itemize}
    \item trimming the lawn, \{outdoor light : on, yard musicplayer : play nature playing music \}
    \item unwinding after work, \{bedroom light : soft, bedroom musicplayer : play chill music, diffuser : on\}
    \item video call with friends, \{living room light : bright, living room smart speaker : video call mode\}
    \item watching sunrise in the balcony, \{balcony light : dim, outdoor speaker : play morning raga, coffee maker : on\}
    \item relaxing with a book, \{living room light : dim, fireplace : low, reading lamp : on\}
\end{itemize}

The user prompts and actions encompass a variety of scenarios, and we prioritize brevity in both user prompts and actions. On average, each user prompt corresponds to 3.1 actions. Following this, we perform a random partition of the data into training, calibration, and evaluation sets, adhering to a ratio of 10:1:2.

The training data serve the purpose of training our agency model and point-wise dependency neural estimator. Calibration data play a crucial role in determining the threshold within conformal prediction, providing a statistical guarantee for our step-by-step planning decision-making process. Lastly, the evaluation data are utilized to assess the performance of our decision-making AI agent.

\subsection{Model Training : Decision-Making Agent}
\label{subsec:model_smart_home_agent}

Our objective is to train a decision-making agent capable of generating plausible actions based on a user prompt and previously executed actions. To accomplish this, we undertake instruction fine-tuning utilizing large language models (LLMs). In our case, we choose Mistral-Instruction-7B-v0.1~\cite{jiang2023mistral} as the LLM, although any proprietary or open-sourced LLMs can be employed for this purpose. It's crucial to note that we do not use the logits outputs from Mistral-Instruction-7B-v0.1, treating it as a black-box LLM. In the following explanation, we will refer to the example in Figure~\ref{fig:illus}, where the user prompt is  {\em water the plants}, and the corresponding actions are \{{\em outdoor lights: on, outdoor speaker: play laid-back music, smart sprinkler: on}\}.

{\bf Instruction Fine-tuning between User Prompt and Action.}  For the instruction, we instruct the home agent to function as a home assistant, deploying actions based on an input command from the user:
\begin{itemize}
\item {[INST]} You are a home assistant, and you receive a command {\color{blue} \em water the plants}.  \\ ~ \\ Please deploy your next action: [/INST]
\end{itemize}

For the output, we surround the action with identifiers {\rm $<$ACT$>$} and {\rm $</$ACT$>$}:
\begin{itemize}
\item $<$ACT$>$ {\color{blue} \em outdoor lights:on} $</$ACT$>$
\end{itemize}
Since $<$ACT$>$ {\em outdoor lights:on} $</$ACT$>$ is not the only possible action, the outputs can also be $<$ACT$>$ {\em outdoor speaker: play laid-back music} $</$ACT$>$ or $<$ACT$>$ {\em smart
sprinkler: on} $</$ACT$>$.

{\bf Extension to Step-by-step Instruction Fine-tuning.} The initial fine-tuning, as described above, does not consider executed actions, leading to a gap in performing step-by-step decision planning. To integrate the history of actions, we randomly select some of the corresponding actions with the user prompt and modify the instruction as follows:
\begin{itemize}
\item {[INST]} You are a home assistant, and you receive a command {\color{blue} \em water the plants}.  \\ ~\\ You deployed $<$ACT$>$ {\color{blue} {\em outdoor lights:on}} $</$ACT$>$, $<$ACT$>$ {\color{blue} {\em outdoor speaker: play laid-back music}} $</$ACT$>$. \\ ~ \\ Please deploy your next action: [/INST]
\end{itemize}
And the output for this instruction would be $<$ACT$>$ {\em smart
sprinkler: on} $</$ACT$>$. 

{\bf Inference with Multiple Actions at Once.} Our setting admits auto-regressive training (i.e., Mistral-Instruct-7B-0.1 is a decoder-only LLM), meaning that next tokens are decoded based on history tokens. In the step-by-step instruction fine-tuning described above, the history actions are decoded one by one. This results in an inference time where multiple actions can be generated at once.

For example, when user prompt is {\em take a hot bath}, a single inference from our decision-making agent generates
\begin{itemize}
\item $<$ACT$>$ {\color{blue} \em smart tubs : fill with hot water} $</$ACT$>$, $<$ACT$>$ {\color{blue} \em bathroom speaker : play relaxing music} $</$ACT$>$, $<$ACT$>$ {\color{blue} \em towel warmer : on} $</$ACT$>$, $<$ACT$>$ {\color{blue} \em bathroom light : soft} $</$ACT$>$, $<$ACT$>$ {\color{blue} \em blinds : down} $</$ACT$>$
\end{itemize}

In contrast, prior approaches on decision planning~\cite{ren2023robots,brohan2023rt,driess2023palm,li2022pre} generate only one single action per inference, whereas our model is designed to generate all actions in a single inference.

\input{fig_tex/dist}
\input{fig_tex/conformal}

\subsection{Model Training : Point-wise Dependency Neural Estimator} 
\label{subsec:model_PD}

We train the objective~\eqref{eq:rpc2} on $\mathcal{D}_{\rm train}$ for $3$ epochs to obtain the estimated point-wise dependency (EPD). In the following, we will delve into the distributions associated with the EPD and elaborate on our utilization of conformal prediction~\cite{shafer2008tutorial,ren2023robots} to derive the threshold for the EPD.

{\bf Distributions for Estimated Point-wise Dependency}. We visualize the distributions for EPD on  $\mathcal{D}_{\rm train}$, $\mathcal{D}_{\rm calib}$, $\mathcal{D}_{\rm eval}$ in Figure~\ref{fig:dist}. Notably, we emphasize the value $1$ in the plots, indicating independence between the current action and the user prompt/taken actions.

Firstly, given the use of neural networks for direct point-wise dependency estimation, the presence of negative EPD values (albeit with low probability) is inevitable.

Secondly, a substantial number of EPD values cluster around the value $1$. Examples include \{``{\em A day of painting}'' and ``{\em living room light : bright}''\} and \{``{\em A quiet evening with a book}'' and ``{\em electric kettle : on}''\}. These instances illustrate that our approach effectively identifies cases where the action and user prompt are not strongly dependent. Quantitatively, approximately $12\%$ of the data exhibits EPD values smaller than $1.21$, indicating that $12\%$ of the annotated actions are not strongly dependent on the corresponding user prompt.

Thirdly, we observe a similar distribution plot across the training, calibration, and test data. This consistency suggests that our training process does not overfit to the training data.

{\bf Conformal Prediction.} Next, we employ conformal prediction~\cite{shafer2008tutorial} to establish the threshold for use in test scenarios. As conformal prediction is not the primary focus of this paper, we present a simplified discussion, and interested readers are encouraged to explore the comprehensive study in the prior work~\cite{ren2023robots} . 

Let $x'_{\rm test}$ be a user prompt along with its taken actions, and  $a_{\rm true}$ be the corresponding true action to be executed given $x'_{\rm test}$. Conformal prediction generates a prediction set $\mathcal{C}(x'_{\rm test})$ with a high probability of containing $a_{\rm true}$:
$$
\mathbb{P}\Big(a_{\rm true} \in \mathcal{C}(x'_{\rm test})\Big) \geq 1 - \epsilon ,
$$
where $\epsilon$ is user-defined (in our case, we adopt $\epsilon=20\%$). To determine $\mathcal{C}(x'_{\rm test})$, we define the non-conformity score as  $50$ minus the EPD values. Subsequently, we identify the $1-\epsilon$ empirical quantile of the non-conformity scores on calibration data,  which is $1.627$ in our case. Finally, we determine $\mathcal{C}(x'_{\rm test}) = \{a\in A | r_\theta(a, x'_{\rm test}) \geq 1.627\}$. 

This discussion indicates that, given $x'_{\rm test}$, conformal prediction provides a statistical guarantee that the probability of the true action lying in the generated actions with scores greater than $1.627$ is greater than $80\%$.

\subsection{Deployment}

Now, we are prepared to deploy our decision-making agent and point-wise dependency neural estimator. For an illustration, please refer to Figure~\ref{fig:illus}. We will use the example in the figure for the following explanation. 

{\bf Greeting and User Prompting.} The initial stage of the deployment kicks off with a greeting from the decision-making agent, followed by the user providing a prompt. In the example, the user prompt is 
\begin{itemize}
    \item [\tiny {[}User{]}] I am exhausted today
\end{itemize}

{\bf Actions Generation and User Interaction.} In the next stage, the trained decision-making home agent proceeds to execute the user prompt and generates a series of actions all at once using a single inference (refer to Section~\ref{subsec:model_smart_home_agent}). In the example, the decision-making agent generates the following actions:
\begin{itemize}
    \item  living room light : dim, musicplayer : play soft sounds, air purifier : on, smart blinds : close, living room musicplayer: play nature sound, coffee maker : on, kitchen light : on
\end{itemize}

Upon closer inspection, it's evident that some actions, such as {\em air purifier : on} and {\em kitchen light: on}, may not be perfect matches for the user prompts. Therefore, we rely on the trained point-wise dependency neural estimator to assist in eliminating these options. By applying the threshold determined using conformal prediction (refer to Section~\ref{subsec:model_PD}), the agent 1) selectively chooses actions with scores above the threshold and 2) asks the user the select one of them: 
\begin{itemize}
    \item [\tiny {[}Agent{]}] Action 1 is (living room light : dim) with point-wise dependency 2.206, \\
    Action 2 is (musicplayer: play soft sounds) with point-wise dependency 1.928, \\
    Action 3 is (living room light: dim) with point-wise dependency 1.837. \\
    Please select an action.
\end{itemize}

In response, the user will select one of them:
\begin{itemize}
    \item [\tiny {[}User{]}] Action 2
\end{itemize}

If there is only one generated action with an estimated point-wise dependency larger than the threshold, the agent will automatically select this option. The interactive process between the agent and the user concludes when no generated actions have an estimated point-wise dependency above the threshold. 

{\bf Actions Conclusion.} In the final stage, all the selected actions will be gathered and executed by the smart home, one by one:
\begin{itemize}
    \item [\tiny {[}Home{]}] 1. musicplayer: play soft sounds \\
    2. living room light: warm \\
    3. living room AC: set temperature 20C \\
    4. aroma diffuser: on
\end{itemize}

%% file: fig_tex/dist.tex
\begin{figure*}[th!]
    \centering
    \includegraphics[width=1.0\linewidth]{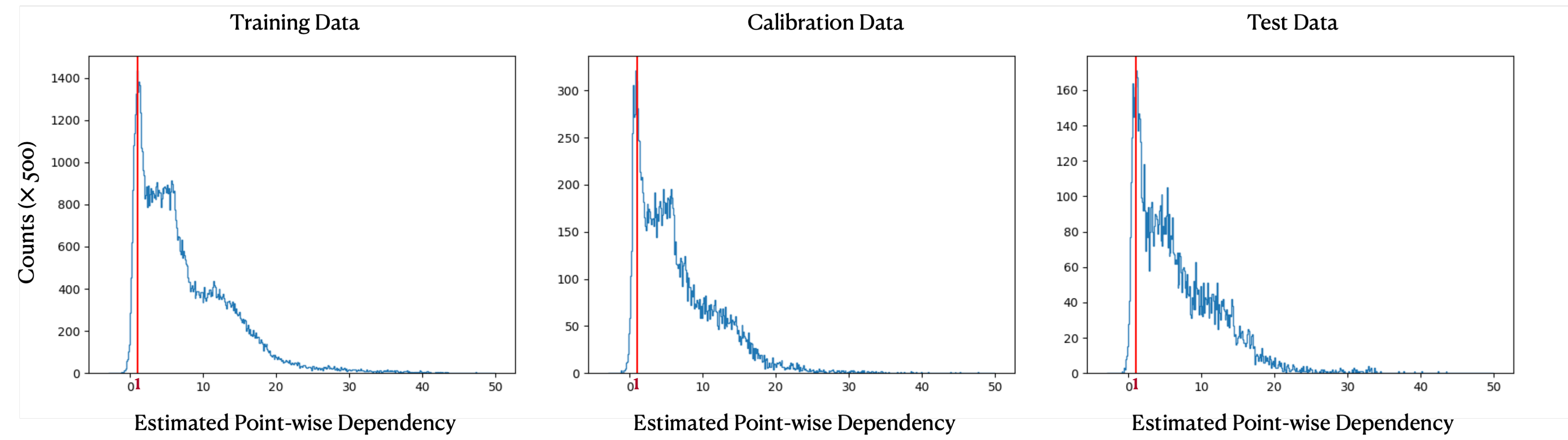}
    \vspace{-6mm}
    \caption{Distributions of estimated point-wise dependency between user prompt, taken actions, and current action.}
    \label{fig:dist}
\end{figure*}

%% file: fig_tex/conformal.tex
\begin{figure}[t!]
    \centering
    \includegraphics[width=0.9\linewidth]{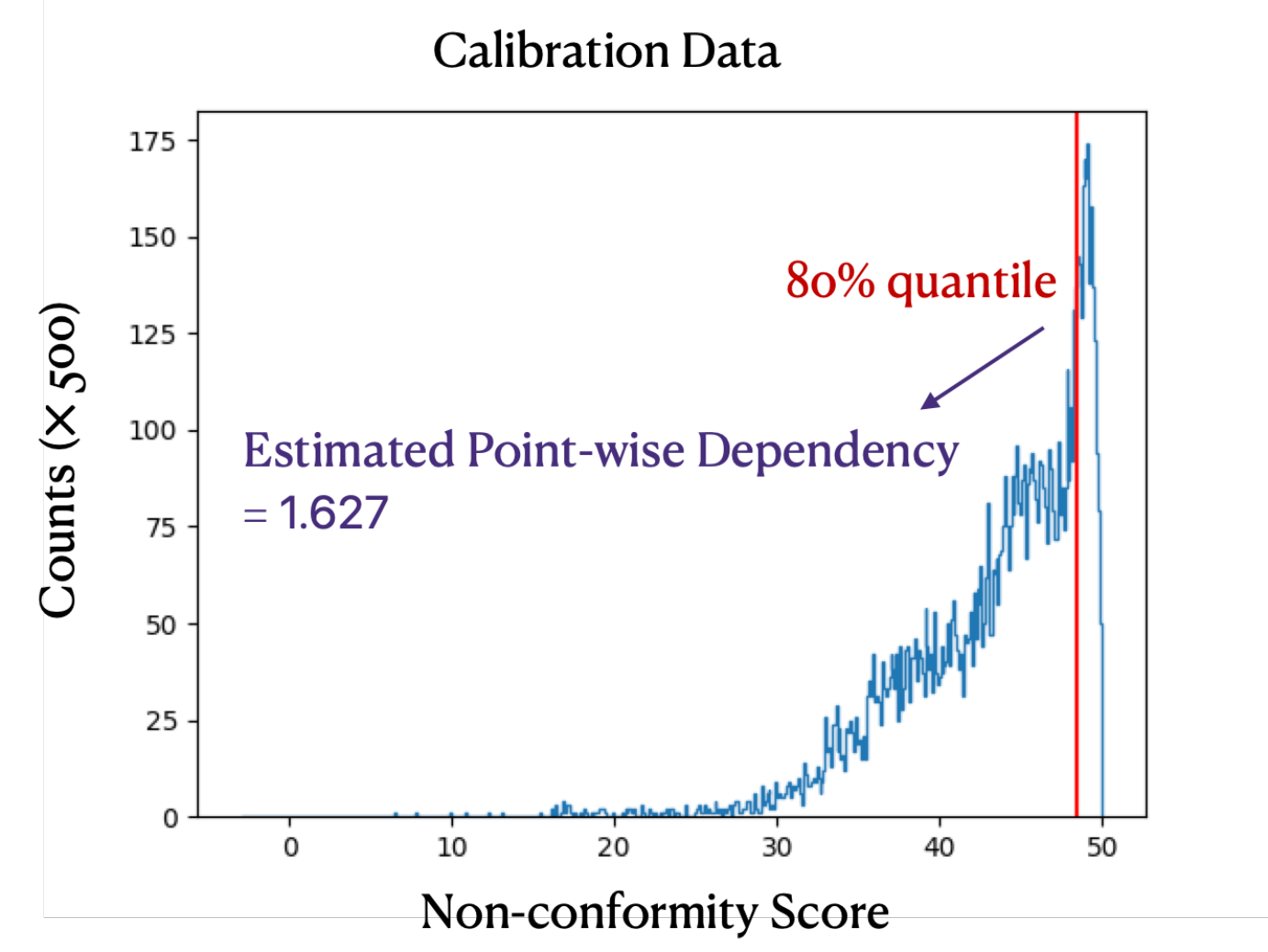}
    \vspace{-3mm}
    \caption{Conformal prediction on calibration data. We define the non-conformity score as $50$ minus the estimated point-wise dependency. Conformal prediction identifies the $80\%$ quantile value to establish the threshold at $1.627$. Consequently, during test scenarios, we secure a statistical guarantee that the probability of the ``true action lying in the generated actions with scores greater than $1.627$'' is greater than $80\%$.}
    \label{fig:conformal}
\end{figure}

%% file: sections/exp.tex
\section{Evaluation}
\label{sec:exp}

This section assesses the performance of our decision-making agent and the point-wise dependency estimator, addressing the following questions:

\begin{itemize}
\item [Q1.] Does step-by-step planning outperform all-at-once action generation? 
\item [Q2.] What is the optimal action generation strategy without human intervention?
\item [Q3.] How does the threshold of the point-wise dependency estimator impact performance?
\end{itemize}

{\bf Metrics.} We evaluate our model on the evaluation data $\mathcal{D}_{\rm eval}$ by assessing the match between generated actions and true annotated actions using an {\em exact match} criterion, which is stringent. For example, {\em musicplayer: play soft sounds} and {\em musicplayer: play soft music} are considered distinct in our evaluation. Our metrics include mean precision, mean recall, and the corresponding F1 score:
\begin{itemize}
\item ${\rm Precision} = \frac{\rm \# \,\,of\,\,the\,\,generated\,\,actions\,\,that\,\,are\,\,true\,\,actions}{\rm \# \,\,of\,\,the\,\,generated\,\,actions}$
\item ${\rm Recall} = \frac{\rm \# \,\,of\,\,the\,\,generated\,\,actions\,\,that\,\,are\,\,true\,\,actions}{\rm \# \,\,of\,\,the\,\,true\,\,actions}$
\item ${\rm F1 \,\, Score} = 2\cdot \frac{\rm precision \cdot recall}{\rm precision + recall}$
\end{itemize}
Precision and recall scores are computed for each user prompt, and the average over all user prompts yields mean precision and mean recall. The F1 score is then calculated based on the mean precision and mean recall. Mean precision assesses the likelihood that a generated action is the true action. Mean recall gauges the likelihood that true actions are encompassed by the generated actions. The F1 score represents the harmonic average of precision and recall.

{\bf Step-by-step Planning vs All-at-Once Generation.} We compare actions generated through step-by-step planning with those generated all-at-once (refer to Inference with Multiple Actions at Once in Section~\ref{subsec:model_smart_home_agent}). Results are presented in Table~\ref{tbl:comparison}. For actions generated all-at-once, we collect them by providing only the user prompt without history actions to the decision-making agent. Regarding step-by-step planning, we consider two scenarios: 1) randomly selecting one action from the suggested actions and 2) always selecting the action with the highest estimated point-wise dependency. All approaches share a common threshold for point-wise dependency set at $1.0$, meaning actions with scores below this threshold will not be generated.

Table results show that actions generated through step-by-step planning outperform those generated all-at-once. We posit that this is because, in step-by-step planning, an action is generated not solely based on the user prompt but also leverages additional information: executed actions. This observation parallels the chain-of-thought process in large language models~\cite{openai2023gpt4}, suggesting that our decision-making agent achieves better performance by making decisions incrementally rather than directly making the final decision.

\input{tbl_tex/step_by_step_vs_all_at_once.tex}

{\bf Best Action Generation Strategy without Human in the Loop.} Observing Table~\ref{tbl:comparison}, we note that employing step-by-step planning and consistently selecting actions with the highest estimated point-wise dependency yields the best performance. The fact that random selection performs worse than selecting actions with the highest point-wise dependency implies that higher estimated point-wise dependency indicates a higher likelihood of being the true action. In future work, we aim to conduct a human study wherein actions are selected one by one by humans, providing a more comprehensive analysis.

{\bf Threshold in the Point-wise Dependency Neural Estimator.} Now, we explore how the threshold $t$ for the point-wise dependency neural estimator affects the overall performance. Actions below the threshold will be discarded during the generation process. We examine three thresholds: $t=0.0$, $t=1.0$, and $t=1.627$. A threshold of $0.0$ implies considering all actions with either negative or positive point-wise dependencies. A threshold of $1.0$ means considering only actions with positive point-wise dependencies. The threshold of $1.627$ is determined during the conformal prediction process in Section~\ref{subsec:model_PD} and is used to obtain a statistical guarantee that the probability of the true action lying in the generated actions is greater than $80\%$. We provide the results in 
Table~\ref{tbl:threshold}.

\input{tbl_tex/threshold.tex} 

Our initial observation indicates that the statistical guarantee obtained from conformal prediction~\cite{ren2023robots,shafer2008tutorial} does not strongly correlate with the actual performance in our evaluation. We attribute this to the strict criterion of {\em exact match} used between generated actions and true actions. Future extensions to our evaluation will include measures of {\em semantic similarity} and {\em human study} in addition to {\em exact match}.

Our second observation indicates that increasing the threshold leads to an improved F1 score. As the threshold rises, mean precision increases while mean recall decreases. This observation is expected, as a higher threshold results in discarding more generated actions, reducing recall. The increase in mean precision suggests that our estimated point-wise dependency serves as an effective selector for actions more likely to be true actions.

%% file: tbl_tex/step_by_step_vs_all_at_once.tex
\begin{table}[t!]
    \caption{Comparison between all-at-once action generation and actions generated through the step-by-step planning process (randomly selecting one action from the suggested actions v.s. always selecting the action with the highest estimated point-wise dependency). The threshold for point-wise dependency is set to $1.0$.}
    \centering
    \small
    \vspace{-1mm}
    \begin{tabular}{cccc}
    \toprule
    \multirow{2}{*}{metrics} & \multirow{2}{*}{all-at-once}  & step-by-step & step-by-step \\
     & & (random) & (maximum) \\
    \midrule
    mean precision & 0.108 & 0.126 & 0.134 \\
    mean recall & 0.200 & 0.193 & 0.212 \\
    F1 score & 0.140 & 0.152 & 0.164 \\
    \bottomrule
    \end{tabular}
    \vspace{-1mm}
    \label{tbl:comparison}
\end{table}

%% file: tbl_tex/threshold.tex
\begin{table}[t!]
    \caption{Analysis of the effects of threshold $t$ for estimated point-wise dependency. 
    Actions below the threshold will be discarded during the generation process.
    }
    \centering
    \small
    \vspace{-1mm}
    \begin{tabular}{cccc}
    \toprule
     \multicolumn{4}{c}{\it all-at-once action generation } \\
     \midrule
     \midrule
    metrics & t = 0.0 & t = 1.0 & t = 1.627 \\
    \midrule
    mean precision & 0.083 & 0.108 & 0.127 \\
    mean recall & 0.235 & 0.200 & 0.167 \\
    F1 score & 0.127 & 0.140 & 0.144 \\
    \midrule
    \midrule
     \multicolumn{4}{c}{\it step-by-step planning (maximum) } \\
     \midrule
     \midrule
    metrics & t = 0.0 & t = 1.0 & t = 1.627 \\
    \midrule
    mean precision & 0.123 & 0.134 & 0.139 \\
    mean recall & 0.232 & 0.212 & 0.209 \\
    F1 score & 0.162 & 0.164 & 0.167 \\
    \bottomrule
    \end{tabular}
    \vspace{-3mm}
    \label{tbl:threshold}
\end{table}

%% file: sections/conclusion.tex
\section{Conclusion}
\label{sec:conclu}

This work explores the design of the AI agent's ability to make decisions based on human requests in natural language. It delves into the challenges of uncertainty quantification in decision-making and proposes an efficient approach using point-wise dependency neural estimation for black-box large language models, opening up the possibility of using powerful proprietary language models. The work further advances the application of this approach in building a decision-making agent capable of step-by-step decision planning, detailing the stages of data collection, model training, and deployment. 

%% file: main.bbl
\begin{thebibliography}{59}
\providecommand{\natexlab}[1]{#1}
\providecommand{\url}[1]{\texttt{#1}}
\expandafter\ifx\csname urlstyle\endcsname\relax
  \providecommand{\doi}[1]{doi: #1}\else
  \providecommand{\doi}{doi: \begingroup \urlstyle{rm}\Url}\fi

\bibitem[Ahn et~al.(2022)Ahn, Brohan, Brown, Chebotar, Cortes, David, Finn, Fu,
  Gopalakrishnan, Hausman, et~al.]{ahn2022can}
Ahn, M., Brohan, A., Brown, N., Chebotar, Y., Cortes, O., David, B., Finn, C.,
  Fu, C., Gopalakrishnan, K., Hausman, K., et~al.
\newblock Do as i can, not as i say: Grounding language in robotic affordances.
\newblock \emph{arXiv preprint arXiv:2204.01691}, 2022.

\bibitem[Aky{\"u}rek et~al.(2023)Aky{\"u}rek, Aky{\"u}rek, Madaan, Kalyan,
  Clark, Wijaya, and Tandon]{akyurek2023rl4f}
Aky{\"u}rek, A.~F., Aky{\"u}rek, E., Madaan, A., Kalyan, A., Clark, P., Wijaya,
  D., and Tandon, N.
\newblock Rl4f: Generating natural language feedback with reinforcement
  learning for repairing model outputs.
\newblock \emph{arXiv preprint arXiv:2305.08844}, 2023.

\bibitem[Alizadeh et~al.(2023)Alizadeh, Mirzadeh, Belenko, Khatamifard, Cho,
  Del~Mundo, Rastegari, and Farajtabar]{alizadeh2023llm}
Alizadeh, K., Mirzadeh, I., Belenko, D., Khatamifard, K., Cho, M., Del~Mundo,
  C.~C., Rastegari, M., and Farajtabar, M.
\newblock Llm in a flash: Efficient large language model inference with limited
  memory.
\newblock \emph{arXiv preprint arXiv:2312.11514}, 2023.

\bibitem[Belghazi et~al.(2018)Belghazi, Baratin, Rajeshwar, Ozair, Bengio,
  Courville, and Hjelm]{belghazi2018mutual}
Belghazi, M.~I., Baratin, A., Rajeshwar, S., Ozair, S., Bengio, Y., Courville,
  A., and Hjelm, D.
\newblock Mutual information neural estimation.
\newblock In \emph{International conference on machine learning}, pp.\
  531--540. PMLR, 2018.

\bibitem[Boiko et~al.(2023)Boiko, MacKnight, and Gomes]{boiko2023emergent}
Boiko, D.~A., MacKnight, R., and Gomes, G.
\newblock Emergent autonomous scientific research capabilities of large
  language models.
\newblock \emph{arXiv preprint arXiv:2304.05332}, 2023.

\bibitem[Brohan et~al.(2023)Brohan, Brown, Carbajal, Chebotar, Chen,
  Choromanski, Ding, Driess, Dubey, Finn, et~al.]{brohan2023rt}
Brohan, A., Brown, N., Carbajal, J., Chebotar, Y., Chen, X., Choromanski, K.,
  Ding, T., Driess, D., Dubey, A., Finn, C., et~al.
\newblock Rt-2: Vision-language-action models transfer web knowledge to robotic
  control.
\newblock \emph{arXiv preprint arXiv:2307.15818}, 2023.

\bibitem[Chen et~al.(2021)Chen, Lu, Rajeswaran, Lee, Grover, Laskin, Abbeel,
  Srinivas, and Mordatch]{chen2021decision}
Chen, L., Lu, K., Rajeswaran, A., Lee, K., Grover, A., Laskin, M., Abbeel, P.,
  Srinivas, A., and Mordatch, I.
\newblock Decision transformer: Reinforcement learning via sequence modeling.
\newblock \emph{Advances in neural information processing systems},
  34:\penalty0 15084--15097, 2021.

\bibitem[Cho et~al.(2014)Cho, Van~Merri{\"e}nboer, Gulcehre, Bahdanau,
  Bougares, Schwenk, and Bengio]{cho2014learning}
Cho, K., Van~Merri{\"e}nboer, B., Gulcehre, C., Bahdanau, D., Bougares, F.,
  Schwenk, H., and Bengio, Y.
\newblock Learning phrase representations using rnn encoder-decoder for
  statistical machine translation.
\newblock \emph{arXiv preprint arXiv:1406.1078}, 2014.

\bibitem[Dai et~al.(2023)Dai, Li, Liu, Zhao, Wu, Song, Shen, Zhu, Li, Li,
  et~al.]{dai2023ad}
Dai, H., Li, Y., Liu, Z., Zhao, L., Wu, Z., Song, S., Shen, Y., Zhu, D., Li,
  X., Li, S., et~al.
\newblock Ad-autogpt: An autonomous gpt for alzheimer's disease infodemiology.
\newblock \emph{arXiv preprint arXiv:2306.10095}, 2023.

\bibitem[Driess et~al.(2023)Driess, Xia, Sajjadi, Lynch, Chowdhery, Ichter,
  Wahid, Tompson, Vuong, Yu, et~al.]{driess2023palm}
Driess, D., Xia, F., Sajjadi, M.~S., Lynch, C., Chowdhery, A., Ichter, B.,
  Wahid, A., Tompson, J., Vuong, Q., Yu, T., et~al.
\newblock Palm-e: An embodied multimodal language model.
\newblock \emph{arXiv preprint arXiv:2303.03378}, 2023.

\bibitem[Fomicheva et~al.(2020)Fomicheva, Sun, Yankovskaya, Blain, Guzm{\'a}n,
  Fishel, Aletras, Chaudhary, and Specia]{fomicheva2020unsupervised}
Fomicheva, M., Sun, S., Yankovskaya, L., Blain, F., Guzm{\'a}n, F., Fishel, M.,
  Aletras, N., Chaudhary, V., and Specia, L.
\newblock Unsupervised quality estimation for neural machine translation.
\newblock \emph{Transactions of the Association for Computational Linguistics},
  8:\penalty0 539--555, 2020.

\bibitem[Guu et~al.(2020)Guu, Lee, Tung, Pasupat, and Chang]{guu2020retrieval}
Guu, K., Lee, K., Tung, Z., Pasupat, P., and Chang, M.
\newblock Retrieval augmented language model pre-training.
\newblock In \emph{International conference on machine learning}, pp.\
  3929--3938. PMLR, 2020.

\bibitem[Hong et~al.(2023)Hong, Zheng, Chen, Cheng, Wang, Zhang, Wang, Yau,
  Lin, Zhou, et~al.]{hong2023metagpt}
Hong, S., Zheng, X., Chen, J., Cheng, Y., Wang, J., Zhang, C., Wang, Z., Yau,
  S. K.~S., Lin, Z., Zhou, L., et~al.
\newblock Metagpt: Meta programming for multi-agent collaborative framework.
\newblock \emph{arXiv preprint arXiv:2308.00352}, 2023.

\bibitem[Huang et~al.(2022)Huang, Xia, Xiao, Chan, Liang, Florence, Zeng,
  Tompson, Mordatch, Chebotar, et~al.]{huang2022inner}
Huang, W., Xia, F., Xiao, T., Chan, H., Liang, J., Florence, P., Zeng, A.,
  Tompson, J., Mordatch, I., Chebotar, Y., et~al.
\newblock Inner monologue: Embodied reasoning through planning with language
  models.
\newblock \emph{arXiv preprint arXiv:2207.05608}, 2022.

\bibitem[Izacard et~al.(2022)Izacard, Lewis, Lomeli, Hosseini, Petroni, Schick,
  Dwivedi-Yu, Joulin, Riedel, and Grave]{izacard2022few}
Izacard, G., Lewis, P., Lomeli, M., Hosseini, L., Petroni, F., Schick, T.,
  Dwivedi-Yu, J., Joulin, A., Riedel, S., and Grave, E.
\newblock Few-shot learning with retrieval augmented language models.
\newblock \emph{arXiv preprint arXiv:2208.03299}, 2022.

\bibitem[Jiang et~al.(2023)Jiang, Sablayrolles, Mensch, Bamford, Chaplot,
  Casas, Bressand, Lengyel, Lample, Saulnier, et~al.]{jiang2023mistral}
Jiang, A.~Q., Sablayrolles, A., Mensch, A., Bamford, C., Chaplot, D.~S., Casas,
  D. d.~l., Bressand, F., Lengyel, G., Lample, G., Saulnier, L., et~al.
\newblock Mistral 7b.
\newblock \emph{arXiv preprint arXiv:2310.06825}, 2023.

\bibitem[Jiang et~al.(2024)Jiang, Sablayrolles, Roux, Mensch, Savary, Bamford,
  Chaplot, de~las Casas, Hanna, Bressand, Lengyel, Bour, Lample, Lavaud,
  Saulnier, Lachaux, Stock, Subramanian, Yang, Antoniak, Scao, Gervet, Lavril,
  Wang, Lacroix, and Sayed]{jiang2024mixtral}
Jiang, A.~Q., Sablayrolles, A., Roux, A., Mensch, A., Savary, B., Bamford, C.,
  Chaplot, D.~S., de~las Casas, D., Hanna, E.~B., Bressand, F., Lengyel, G.,
  Bour, G., Lample, G., Lavaud, L.~R., Saulnier, L., Lachaux, M.-A., Stock, P.,
  Subramanian, S., Yang, S., Antoniak, S., Scao, T.~L., Gervet, T., Lavril, T.,
  Wang, T., Lacroix, T., and Sayed, W.~E.
\newblock Mixtral of experts, 2024.

\bibitem[Kadavath et~al.(2022)Kadavath, Conerly, Askell, Henighan, Drain,
  Perez, Schiefer, Hatfield-Dodds, DasSarma, Tran-Johnson,
  et~al.]{kadavath2022language}
Kadavath, S., Conerly, T., Askell, A., Henighan, T., Drain, D., Perez, E.,
  Schiefer, N., Hatfield-Dodds, Z., DasSarma, N., Tran-Johnson, E., et~al.
\newblock Language models (mostly) know what they know.
\newblock \emph{arXiv preprint arXiv:2207.05221}, 2022.

\bibitem[Kuhn et~al.(2023)Kuhn, Gal, and Farquhar]{kuhn2023semantic}
Kuhn, L., Gal, Y., and Farquhar, S.
\newblock Semantic uncertainty: Linguistic invariances for uncertainty
  estimation in natural language generation.
\newblock \emph{arXiv preprint arXiv:2302.09664}, 2023.

\bibitem[Kumar et~al.(2023)Kumar, Lu, Gupta, Palepu, Bellamy, Raskar, and
  Beam]{kumar2023conformal}
Kumar, B., Lu, C., Gupta, G., Palepu, A., Bellamy, D., Raskar, R., and Beam, A.
\newblock Conformal prediction with large language models for multi-choice
  question answering.
\newblock \emph{arXiv preprint arXiv:2305.18404}, 2023.

\bibitem[Lee et~al.(2018)Lee, Lee, Lee, and Shin]{lee2018simple}
Lee, K., Lee, K., Lee, H., and Shin, J.
\newblock A simple unified framework for detecting out-of-distribution samples
  and adversarial attacks.
\newblock \emph{Advances in neural information processing systems}, 31, 2018.

\bibitem[Li et~al.(2023)Li, Hammoud, Itani, Khizbullin, and
  Ghanem]{li2023camel}
Li, G., Hammoud, H. A. A.~K., Itani, H., Khizbullin, D., and Ghanem, B.
\newblock Camel: Communicative agents for" mind" exploration of large scale
  language model society.
\newblock \emph{arXiv preprint arXiv:2303.17760}, 2023.

\bibitem[Li et~al.(2022)Li, Puig, Paxton, Du, Wang, Fan, Chen, Huang,
  Aky{\"u}rek, Anandkumar, et~al.]{li2022pre}
Li, S., Puig, X., Paxton, C., Du, Y., Wang, C., Fan, L., Chen, T., Huang,
  D.-A., Aky{\"u}rek, E., Anandkumar, A., et~al.
\newblock Pre-trained language models for interactive decision-making.
\newblock \emph{Advances in Neural Information Processing Systems},
  35:\penalty0 31199--31212, 2022.

\bibitem[Lin et~al.(2023)Lin, Trivedi, and Sun]{lin2023generating}
Lin, Z., Trivedi, S., and Sun, J.
\newblock Generating with confidence: Uncertainty quantification for black-box
  large language models.
\newblock \emph{arXiv preprint arXiv:2305.19187}, 2023.

\bibitem[Liu et~al.(2023{\natexlab{a}})Liu, Sferrazza, and
  Abbeel]{liu2023languages}
Liu, H., Sferrazza, C., and Abbeel, P.
\newblock Languages are rewards: Hindsight finetuning using human feedback.
\newblock \emph{arXiv preprint arXiv:2302.02676}, 2023{\natexlab{a}}.

\bibitem[Liu et~al.(2023{\natexlab{b}})Liu, Yang, Jia, Zhang, Zhou, Dai, Yang,
  and Vosoughi]{liu2023training}
Liu, R., Yang, R., Jia, C., Zhang, G., Zhou, D., Dai, A.~M., Yang, D., and
  Vosoughi, S.
\newblock Training socially aligned language models in simulated human society.
\newblock \emph{arXiv preprint arXiv:2305.16960}, 2023{\natexlab{b}}.

\bibitem[Lu et~al.(2023)Lu, Peng, Cheng, Galley, Chang, Wu, Zhu, and
  Gao]{lu2023chameleon}
Lu, P., Peng, B., Cheng, H., Galley, M., Chang, K.-W., Wu, Y.~N., Zhu, S.-C.,
  and Gao, J.
\newblock Chameleon: Plug-and-play compositional reasoning with large language
  models.
\newblock \emph{arXiv preprint arXiv:2304.09842}, 2023.

\bibitem[Malinin \& Gales(2020)Malinin and Gales]{malinin2020uncertainty}
Malinin, A. and Gales, M.
\newblock Uncertainty estimation in autoregressive structured prediction.
\newblock \emph{arXiv preprint arXiv:2002.07650}, 2020.

\bibitem[Nakano et~al.(2021)Nakano, Hilton, Balaji, Wu, Ouyang, Kim, Hesse,
  Jain, Kosaraju, Saunders, et~al.]{nakano2021webgpt}
Nakano, R., Hilton, J., Balaji, S., Wu, J., Ouyang, L., Kim, C., Hesse, C.,
  Jain, S., Kosaraju, V., Saunders, W., et~al.
\newblock Webgpt: Browser-assisted question-answering with human feedback.
\newblock \emph{arXiv preprint arXiv:2112.09332}, 2021.

\bibitem[OpenAI et~al.(2023)OpenAI, :, Achiam, Adler, Agarwal, Ahmad, Akkaya,
  Aleman, Almeida, Altenschmidt, Altman, Anadkat, Avila, Babuschkin, Balaji,
  Balcom, Baltescu, Bao, Bavarian, Belgum, Bello, Berdine, Bernadett-Shapiro,
  Berner, Bogdonoff, Boiko, Boyd, Brakman, Brockman, Brooks, Brundage, Button,
  Cai, Campbell, Cann, Carey, Carlson, Carmichael, Chan, Chang, Chantzis, Chen,
  Chen, Chen, Chen, Chen, Chess, Cho, Chu, Chung, Cummings, Currier, Dai,
  Decareaux, Degry, Deutsch, Deville, Dhar, Dohan, Dowling, Dunning, Ecoffet,
  Eleti, Eloundou, Farhi, Fedus, Felix, Fishman, Forte, Fulford, Gao, Georges,
  Gibson, Goel, Gogineni, Goh, Gontijo-Lopes, Gordon, Grafstein, Gray, Greene,
  Gross, Gu, Guo, Hallacy, Han, Harris, He, Heaton, Heidecke, Hesse, Hickey,
  Hickey, Hoeschele, Houghton, Hsu, Hu, Hu, Huizinga, Jain, Jain, Jang, Jiang,
  Jiang, Jin, Jin, Jomoto, Jonn, Jun, Kaftan, Łukasz Kaiser, Kamali,
  Kanitscheider, Keskar, Khan, Kilpatrick, Kim, Kim, Kim, Kirchner, Kiros,
  Knight, Kokotajlo, Łukasz Kondraciuk, Kondrich, Konstantinidis, Kosic,
  Krueger, Kuo, Lampe, Lan, Lee, Leike, Leung, Levy, Li, Lim, Lin, Lin, Litwin,
  Lopez, Lowe, Lue, Makanju, Malfacini, Manning, Markov, Markovski, Martin,
  Mayer, Mayne, McGrew, McKinney, McLeavey, McMillan, McNeil, Medina, Mehta,
  Menick, Metz, Mishchenko, Mishkin, Monaco, Morikawa, Mossing, Mu, Murati,
  Murk, Mély, Nair, Nakano, Nayak, Neelakantan, Ngo, Noh, Ouyang, O'Keefe,
  Pachocki, Paino, Palermo, Pantuliano, Parascandolo, Parish, Parparita,
  Passos, Pavlov, Peng, Perelman, de~Avila Belbute~Peres, Petrov,
  de~Oliveira~Pinto, Michael, Pokorny, Pokrass, Pong, Powell, Power, Power,
  Proehl, Puri, Radford, Rae, Ramesh, Raymond, Real, Rimbach, Ross, Rotsted,
  Roussez, Ryder, Saltarelli, Sanders, Santurkar, Sastry, Schmidt, Schnurr,
  Schulman, Selsam, Sheppard, Sherbakov, Shieh, Shoker, Shyam, Sidor, Sigler,
  Simens, Sitkin, Slama, Sohl, Sokolowsky, Song, Staudacher, Such, Summers,
  Sutskever, Tang, Tezak, Thompson, Tillet, Tootoonchian, Tseng, Tuggle,
  Turley, Tworek, Uribe, Vallone, Vijayvergiya, Voss, Wainwright, Wang, Wang,
  Wang, Ward, Wei, Weinmann, Welihinda, Welinder, Weng, Weng, Wiethoff,
  Willner, Winter, Wolrich, Wong, Workman, Wu, Wu, Wu, Xiao, Xu, Yoo, Yu, Yuan,
  Zaremba, Zellers, Zhang, Zhang, Zhao, Zheng, Zhuang, Zhuk, and
  Zoph]{openai2023gpt4}
OpenAI, :, Achiam, J., Adler, S., Agarwal, S., Ahmad, L., Akkaya, I., Aleman,
  F.~L., Almeida, D., Altenschmidt, J., Altman, S., Anadkat, S., Avila, R.,
  Babuschkin, I., Balaji, S., Balcom, V., Baltescu, P., Bao, H., Bavarian, M.,
  Belgum, J., Bello, I., Berdine, J., Bernadett-Shapiro, G., Berner, C.,
  Bogdonoff, L., Boiko, O., Boyd, M., Brakman, A.-L., Brockman, G., Brooks, T.,
  Brundage, M., Button, K., Cai, T., Campbell, R., Cann, A., Carey, B.,
  Carlson, C., Carmichael, R., Chan, B., Chang, C., Chantzis, F., Chen, D.,
  Chen, S., Chen, R., Chen, J., Chen, M., Chess, B., Cho, C., Chu, C., Chung,
  H.~W., Cummings, D., Currier, J., Dai, Y., Decareaux, C., Degry, T., Deutsch,
  N., Deville, D., Dhar, A., Dohan, D., Dowling, S., Dunning, S., Ecoffet, A.,
  Eleti, A., Eloundou, T., Farhi, D., Fedus, L., Felix, N., Fishman, S.~P.,
  Forte, J., Fulford, I., Gao, L., Georges, E., Gibson, C., Goel, V., Gogineni,
  T., Goh, G., Gontijo-Lopes, R., Gordon, J., Grafstein, M., Gray, S., Greene,
  R., Gross, J., Gu, S.~S., Guo, Y., Hallacy, C., Han, J., Harris, J., He, Y.,
  Heaton, M., Heidecke, J., Hesse, C., Hickey, A., Hickey, W., Hoeschele, P.,
  Houghton, B., Hsu, K., Hu, S., Hu, X., Huizinga, J., Jain, S., Jain, S.,
  Jang, J., Jiang, A., Jiang, R., Jin, H., Jin, D., Jomoto, S., Jonn, B., Jun,
  H., Kaftan, T., Łukasz Kaiser, Kamali, A., Kanitscheider, I., Keskar, N.~S.,
  Khan, T., Kilpatrick, L., Kim, J.~W., Kim, C., Kim, Y., Kirchner, H., Kiros,
  J., Knight, M., Kokotajlo, D., Łukasz Kondraciuk, Kondrich, A.,
  Konstantinidis, A., Kosic, K., Krueger, G., Kuo, V., Lampe, M., Lan, I., Lee,
  T., Leike, J., Leung, J., Levy, D., Li, C.~M., Lim, R., Lin, M., Lin, S.,
  Litwin, M., Lopez, T., Lowe, R., Lue, P., Makanju, A., Malfacini, K.,
  Manning, S., Markov, T., Markovski, Y., Martin, B., Mayer, K., Mayne, A.,
  McGrew, B., McKinney, S.~M., McLeavey, C., McMillan, P., McNeil, J., Medina,
  D., Mehta, A., Menick, J., Metz, L., Mishchenko, A., Mishkin, P., Monaco, V.,
  Morikawa, E., Mossing, D., Mu, T., Murati, M., Murk, O., Mély, D., Nair, A.,
  Nakano, R., Nayak, R., Neelakantan, A., Ngo, R., Noh, H., Ouyang, L.,
  O'Keefe, C., Pachocki, J., Paino, A., Palermo, J., Pantuliano, A.,
  Parascandolo, G., Parish, J., Parparita, E., Passos, A., Pavlov, M., Peng,
  A., Perelman, A., de~Avila Belbute~Peres, F., Petrov, M., de~Oliveira~Pinto,
  H.~P., Michael, Pokorny, Pokrass, M., Pong, V., Powell, T., Power, A., Power,
  B., Proehl, E., Puri, R., Radford, A., Rae, J., Ramesh, A., Raymond, C.,
  Real, F., Rimbach, K., Ross, C., Rotsted, B., Roussez, H., Ryder, N.,
  Saltarelli, M., Sanders, T., Santurkar, S., Sastry, G., Schmidt, H., Schnurr,
  D., Schulman, J., Selsam, D., Sheppard, K., Sherbakov, T., Shieh, J., Shoker,
  S., Shyam, P., Sidor, S., Sigler, E., Simens, M., Sitkin, J., Slama, K.,
  Sohl, I., Sokolowsky, B., Song, Y., Staudacher, N., Such, F.~P., Summers, N.,
  Sutskever, I., Tang, J., Tezak, N., Thompson, M., Tillet, P., Tootoonchian,
  A., Tseng, E., Tuggle, P., Turley, N., Tworek, J., Uribe, J. F.~C., Vallone,
  A., Vijayvergiya, A., Voss, C., Wainwright, C., Wang, J.~J., Wang, A., Wang,
  B., Ward, J., Wei, J., Weinmann, C., Welihinda, A., Welinder, P., Weng, J.,
  Weng, L., Wiethoff, M., Willner, D., Winter, C., Wolrich, S., Wong, H.,
  Workman, L., Wu, S., Wu, J., Wu, M., Xiao, K., Xu, T., Yoo, S., Yu, K., Yuan,
  Q., Zaremba, W., Zellers, R., Zhang, C., Zhang, M., Zhao, S., Zheng, T.,
  Zhuang, J., Zhuk, W., and Zoph, B.
\newblock Gpt-4 technical report, 2023.

\bibitem[Park et~al.(2023)Park, O'Brien, Cai, Morris, Liang, and
  Bernstein]{park2023generative}
Park, J.~S., O'Brien, J., Cai, C.~J., Morris, M.~R., Liang, P., and Bernstein,
  M.~S.
\newblock Generative agents: Interactive simulacra of human behavior.
\newblock In \emph{Proceedings of the 36th Annual ACM Symposium on User
  Interface Software and Technology}, pp.\  1--22, 2023.

\bibitem[Peng et~al.(2023)Peng, Galley, He, Cheng, Xie, Hu, Huang, Liden, Yu,
  Chen, et~al.]{peng2023check}
Peng, B., Galley, M., He, P., Cheng, H., Xie, Y., Hu, Y., Huang, Q., Liden, L.,
  Yu, Z., Chen, W., et~al.
\newblock Check your facts and try again: Improving large language models with
  external knowledge and automated feedback.
\newblock \emph{arXiv preprint arXiv:2302.12813}, 2023.

\bibitem[Qian et~al.(2023)Qian, Cong, Yang, Chen, Su, Xu, Liu, and
  Sun]{qian2023communicative}
Qian, C., Cong, X., Yang, C., Chen, W., Su, Y., Xu, J., Liu, Z., and Sun, M.
\newblock Communicative agents for software development.
\newblock \emph{arXiv preprint arXiv:2307.07924}, 2023.

\bibitem[Quach et~al.(2023)Quach, Fisch, Schuster, Yala, Sohn, Jaakkola, and
  Barzilay]{quach2023conformal}
Quach, V., Fisch, A., Schuster, T., Yala, A., Sohn, J.~H., Jaakkola, T.~S., and
  Barzilay, R.
\newblock Conformal language modeling.
\newblock \emph{arXiv preprint arXiv:2306.10193}, 2023.

\bibitem[Ren et~al.(2023)Ren, Dixit, Bodrova, Singh, Tu, Brown, Xu, Takayama,
  Xia, Varley, et~al.]{ren2023robots}
Ren, A.~Z., Dixit, A., Bodrova, A., Singh, S., Tu, S., Brown, N., Xu, P.,
  Takayama, L., Xia, F., Varley, J., et~al.
\newblock Robots that ask for help: Uncertainty alignment for large language
  model planners.
\newblock \emph{arXiv preprint arXiv:2307.01928}, 2023.

\bibitem[Ren et~al.(2022)Ren, Luo, Zhao, Krishna, Saleh, Lakshminarayanan, and
  Liu]{ren2022out}
Ren, J., Luo, J., Zhao, Y., Krishna, K., Saleh, M., Lakshminarayanan, B., and
  Liu, P.~J.
\newblock Out-of-distribution detection and selective generation for
  conditional language models.
\newblock \emph{arXiv preprint arXiv:2209.15558}, 2022.

\bibitem[Russell \& Norvig(2010)Russell and Norvig]{russell2010artificial}
Russell, S.~J. and Norvig, P.
\newblock \emph{Artificial intelligence a modern approach}.
\newblock London, 2010.

\bibitem[Schick et~al.(2023)Schick, Dwivedi-Yu, Dess{\`\i}, Raileanu, Lomeli,
  Zettlemoyer, Cancedda, and Scialom]{schick2023toolformer}
Schick, T., Dwivedi-Yu, J., Dess{\`\i}, R., Raileanu, R., Lomeli, M.,
  Zettlemoyer, L., Cancedda, N., and Scialom, T.
\newblock Toolformer: Language models can teach themselves to use tools.
\newblock \emph{arXiv preprint arXiv:2302.04761}, 2023.

\bibitem[Sha et~al.(2023)Sha, Mu, Jiang, Chen, Xu, Luo, Li, Tomizuka, Zhan, and
  Ding]{sha2023languagempc}
Sha, H., Mu, Y., Jiang, Y., Chen, L., Xu, C., Luo, P., Li, S.~E., Tomizuka, M.,
  Zhan, W., and Ding, M.
\newblock Languagempc: Large language models as decision makers for autonomous
  driving.
\newblock \emph{arXiv preprint arXiv:2310.03026}, 2023.

\bibitem[Shafer \& Vovk(2008)Shafer and Vovk]{shafer2008tutorial}
Shafer, G. and Vovk, V.
\newblock A tutorial on conformal prediction.
\newblock \emph{Journal of Machine Learning Research}, 9\penalty0 (3), 2008.

\bibitem[Smith(2013)]{smith2013uncertainty}
Smith, R.~C.
\newblock \emph{Uncertainty quantification: theory, implementation, and
  applications}, volume~12.
\newblock Siam, 2013.

\bibitem[Sumers et~al.(2023)Sumers, Yao, Narasimhan, and
  Griffiths]{sumers2023cognitive}
Sumers, T.~R., Yao, S., Narasimhan, K., and Griffiths, T.~L.
\newblock Cognitive architectures for language agents.
\newblock \emph{arXiv preprint arXiv:2309.02427}, 2023.

\bibitem[Sun et~al.(2020)Sun, Yu, Song, Liu, Yang, and Zhou]{sun2020mobilebert}
Sun, Z., Yu, H., Song, X., Liu, R., Yang, Y., and Zhou, D.
\newblock Mobilebert: a compact task-agnostic bert for resource-limited
  devices.
\newblock \emph{arXiv preprint arXiv:2004.02984}, 2020.

\bibitem[Takayama \& Arase(2019)Takayama and Arase]{takayama2019relevant}
Takayama, J. and Arase, Y.
\newblock Relevant and informative response generation using pointwise mutual
  information.
\newblock In \emph{Proceedings of the First Workshop on NLP for Conversational
  AI}, pp.\  133--138, 2019.

\bibitem[Thirunavukarasu et~al.(2023)Thirunavukarasu, Ting, Elangovan,
  Gutierrez, Tan, and Ting]{thirunavukarasu2023large}
Thirunavukarasu, A.~J., Ting, D. S.~J., Elangovan, K., Gutierrez, L., Tan,
  T.~F., and Ting, D. S.~W.
\newblock Large language models in medicine.
\newblock \emph{Nature medicine}, 29\penalty0 (8):\penalty0 1930--1940, 2023.

\bibitem[Touvron et~al.(2023{\natexlab{a}})Touvron, Lavril, Izacard, Martinet,
  Lachaux, Lacroix, Rozi{\`e}re, Goyal, Hambro, Azhar,
  et~al.]{touvron2023llama}
Touvron, H., Lavril, T., Izacard, G., Martinet, X., Lachaux, M.-A., Lacroix,
  T., Rozi{\`e}re, B., Goyal, N., Hambro, E., Azhar, F., et~al.
\newblock Llama: Open and efficient foundation language models.
\newblock \emph{arXiv preprint arXiv:2302.13971}, 2023{\natexlab{a}}.

\bibitem[Touvron et~al.(2023{\natexlab{b}})Touvron, Martin, Stone, Albert,
  Almahairi, Babaei, Bashlykov, Batra, Bhargava, Bhosale,
  et~al.]{touvron2023llama2}
Touvron, H., Martin, L., Stone, K., Albert, P., Almahairi, A., Babaei, Y.,
  Bashlykov, N., Batra, S., Bhargava, P., Bhosale, S., et~al.
\newblock Llama 2: Open foundation and fine-tuned chat models.
\newblock \emph{arXiv preprint arXiv:2307.09288}, 2023{\natexlab{b}}.

\bibitem[Tsai et~al.(2020)Tsai, Zhao, Yamada, Morency, and
  Salakhutdinov]{tsai2020neural}
Tsai, Y.-H.~H., Zhao, H., Yamada, M., Morency, L.-P., and Salakhutdinov, R.~R.
\newblock Neural methods for point-wise dependency estimation.
\newblock \emph{Advances in Neural Information Processing Systems},
  33:\penalty0 62--72, 2020.

\bibitem[Tsai et~al.(2021)Tsai, Ma, Yang, Zhao, Morency, and
  Salakhutdinov]{tsai2021self}
Tsai, Y.-H.~H., Ma, M.~Q., Yang, M., Zhao, H., Morency, L.-P., and
  Salakhutdinov, R.
\newblock Self-supervised representation learning with relative predictive
  coding.
\newblock \emph{arXiv preprint arXiv:2103.11275}, 2021.

\bibitem[Tsai et~al.(2023)Tsai, Dhar, Li, Zhang, and Zhang]{tsai2023multimodal}
Tsai, Y.-H.~H., Dhar, V., Li, J., Zhang, B., and Zhang, J.
\newblock Multimodal large language model for visual navigation.
\newblock \emph{arXiv preprint arXiv:2310.08669}, 2023.

\bibitem[van~der Poel et~al.(2022)van~der Poel, Cotterell, and
  Meister]{van2022mutual}
van~der Poel, L., Cotterell, R., and Meister, C.
\newblock Mutual information alleviates hallucinations in abstractive
  summarization.
\newblock \emph{arXiv preprint arXiv:2210.13210}, 2022.

\bibitem[Vazhentsev et~al.(2022)Vazhentsev, Kuzmin, Shelmanov, Tsvigun,
  Tsymbalov, Fedyanin, Panov, Panchenko, Gusev, Burtsev,
  et~al.]{vazhentsev2022uncertainty}
Vazhentsev, A., Kuzmin, G., Shelmanov, A., Tsvigun, A., Tsymbalov, E.,
  Fedyanin, K., Panov, M., Panchenko, A., Gusev, G., Burtsev, M., et~al.
\newblock Uncertainty estimation of transformer predictions for
  misclassification detection.
\newblock In \emph{Proceedings of the 60th Annual Meeting of the Association
  for Computational Linguistics (Volume 1: Long Papers)}, pp.\  8237--8252,
  2022.

\bibitem[Wooldridge \& Jennings(1995)Wooldridge and
  Jennings]{wooldridge1995intelligent}
Wooldridge, M. and Jennings, N.~R.
\newblock Intelligent agents: Theory and practice.
\newblock \emph{The knowledge engineering review}, 10\penalty0 (2):\penalty0
  115--152, 1995.

\bibitem[Wu et~al.(2023)Wu, Bansal, Zhang, Wu, Zhang, Zhu, Li, Jiang, Zhang,
  and Wang]{wu2023autogen}
Wu, Q., Bansal, G., Zhang, J., Wu, Y., Zhang, S., Zhu, E., Li, B., Jiang, L.,
  Zhang, X., and Wang, C.
\newblock Autogen: Enabling next-gen llm applications via multi-agent
  conversation framework.
\newblock \emph{arXiv preprint arXiv:2308.08155}, 2023.

\bibitem[Yang et~al.(2023)Yang, Yue, and He]{yang2023auto}
Yang, H., Yue, S., and He, Y.
\newblock Auto-gpt for online decision making: Benchmarks and additional
  opinions.
\newblock \emph{arXiv preprint arXiv:2306.02224}, 2023.

\bibitem[Yao et~al.(2022)Yao, Zhao, Yu, Du, Shafran, Narasimhan, and
  Cao]{yao2022react}
Yao, S., Zhao, J., Yu, D., Du, N., Shafran, I., Narasimhan, K., and Cao, Y.
\newblock React: Synergizing reasoning and acting in language models.
\newblock \emph{arXiv preprint arXiv:2210.03629}, 2022.

\bibitem[Yoo et~al.(2022)Yoo, Kim, Jang, and Kwak]{yoo2022detection}
Yoo, K., Kim, J., Jang, J., and Kwak, N.
\newblock Detection of word adversarial examples in text classification:
  Benchmark and baseline via robust density estimation.
\newblock \emph{arXiv preprint arXiv:2203.01677}, 2022.

\bibitem[Zhang et~al.(2023)Zhang, Li, Cui, Cai, Liu, Fu, Huang, Zhao, Zhang,
  Chen, et~al.]{zhang2023siren}
Zhang, Y., Li, Y., Cui, L., Cai, D., Liu, L., Fu, T., Huang, X., Zhao, E.,
  Zhang, Y., Chen, Y., et~al.
\newblock Siren's song in the ai ocean: A survey on hallucination in large
  language models.
\newblock \emph{arXiv preprint arXiv:2309.01219}, 2023.

\bibitem[Zhou et~al.(2023)Zhou, Hong, and Wu]{zhou2023navgpt}
Zhou, G., Hong, Y., and Wu, Q.
\newblock Navgpt: Explicit reasoning in vision-and-language navigation with
  large language models.
\newblock \emph{arXiv preprint arXiv:2305.16986}, 2023.

\end{thebibliography}
